%% file: molec_tr.tex
\documentclass[12pt]{article}
\usepackage{ulem}
\usepackage[dvips]{graphics}
\usepackage{epsf}
\title{Text as Statistical Mechanics Object}
\author{K. Koroutchev$^{1}$
and
E.Korutcheva$^{2}$\thanks{Also at Depto. F\'{\i}sica Fundamental, UNED
 c/Senda del Rey No 9, 28080 Madrid, Spain}\\
$^{1}$Escuela Polit\'ecnica Superior,
  Universidad Aut\'onoma de Madrid, \\
  28049 Canto Blanco, Madrid, Spain\\
  $^{2}$G.Nadjakov Inst. Solid State
 Physics, \\
 Bulgarian Academy of Sciences, Sofia, Bulgaria
\\and
\\The Abdus Salam International Center for Theoretical Physics, \\
Trieste, Italy
}
\begin{document}
\date{\empty}
\maketitle
\begin{center}
MIRAMARE, TRIESTE

July, 2007
\end{center}
\input{abstract_tr.tex}
\input{intro_tr.tex}

\input{method_tr.tex}

\input{numerical_tr.tex}

\input{conclusion_tr.tex}

\section*{Acknowledgments:}
The authors thank the financial help from the 
International Center for Theoretical
Physics, Trieste, Italy, where the main part of this work has been
performed. 
Especially we would like to thank the Statistical
Physics group at ICTP for stimulating discussions. The work is partly
supported by Grants TIN 2004-07676-G01-01 (K.K.) and
DGI.M.CyT FIS2005-1729 (E.K.) from the Spanish Ministry of Science
and Education.

\end{document}

%% file: abstract_tr.tex
\begin{abstract}
In this article we present a model of human written text
based on statistical mechanics approach by deriving the
potential energy for different parts of the 
text using large text corpus. 


We have checked the results numerically
and found that the ``specific heat'' parameter 
effectively separates the closed class words 
from the specific terms used in the text.
\end{abstract}

%% file: intro_tr.tex
\section{Introduction}

Let us imagine that we are looking for some article on the WEB.
Probably the first thing we will do is to enter in Google and type
some keywords. If we type a query like 
``I am looking for an article about statistical mechanics of images'',
although it is exactly what we want,
we will probably get nothing related to the subject or we will get only a
content partially related to it. So, it would be better to refine the
query to something like ``image ``statistical mechanics'''' in order to
get some reasonable results. To extract useful words we do not use
the structure of the language -- actually we ignore it. 
Also, we hardly use the common words of the sentence in the query.
What we use are some statistical estimations of 
the parts of the query and words that stick well with the meaning
of the query. 
This leads us to the idea to regard the text as a statistical mixture of its
parts that sticks well with its meaning. 
Of course, the text must stick well with the language in which it is 
written as well. Therefore, we can consider the text as conditioned
to the language in which it is written.
We can even consider that the text must stick well
to the area in which it belongs, 
as for example ``nonlinear physics'' or ``novels of 17$^{\rm th}$
century''.

Google is a product of some 15 years evolution of WEB page search.
What this evolution shows is that if we are 
looking for the meaning of a text, we must look for specific,
statistically salient keywords, that are supposed 
to be present in it, largely ignoring the
syntactic and the semantics structure of the language. 
This gives us the inspiration to build a statistical mechanics model
of a human written text, considering it as composed by its
``particles'' -- the words.

The best way to do the analysis of a text, 
written is some language\footnote{Here we will not consider texts like the
genome's sequence, computer logs, multi-language texts and similar.},
is to have some exact descriptions of
the language, for example, weighted 
context free grammar.  
However, it is not clear if such a description exists, because usually
people do not speak grammatically correctly. Some trivial grammars
  always exist, for example grammars that allow all possible strings 
of the alphabet.
  Due to the fact that these strings do not constrain the expressions, the
  information they carry 
regarding the statistical properties of
  the language, is very poor.
Moreover, having in mind the Zipf's law of the frequency distribution of the
words \cite{zipf}, 
even if 
reasonable grammar exists, 
in a single text of arbitrary
length we will 
have some 40\% halomorphemes.\footnote{Halomorphemes are the words that
  occur only once in 
a text.}
As a consequence, the length of the grammar will be of the order of
the length of the text 
for {\bf any} text we choose.
Therefore, it is easier to consider the language as a set of all the
texts spoken/written 
in that language. 
Using statistical arguments, we do not need all texts,
but only a significantly 
large random set of texts in order to treat the problem.

The model we investigate consists of a text $T$ and a vocabulary $V$,
written in one and the same language. 
The vocabulary is formed using all the words of 
some huge collection of texts, 
written in that language.

A text that treats some well-defined subject is highly restricted by this
subject. 
The language, as a whole, has no such restriction. Therefore, the
relative excess 
(or higher frequency) of a word in the vocabulary is a normal situation. 

On contrary, the relative excess of a word in the text has a specific
meaning. 
If the word 
is with much higher frequency of occurrence in the text than in the common language, that can be 
interpreted as an indication that this text treats exactly a subject expressed by this 
word, e.g. that the word is a {\it specific term} or {\it keyword} in the text. 
This is the first class of words in the text that we will consider in this paper.

On the other hand, the text will always contain words that are common
in the language, 
which have more or less the same frequency in any text and in the vocabulary.
A large fraction of the words of that type will be formed by the so-called {\it function words}. 
These words by themselves carry no meaning, but are essential for expressing the language structure. 
A typical example of a function word in English is the word ``the''. 
A similar and more strictly defined category is the class of closed class
words that, by definition, 
are the words which do not change their form in any text.

Finally, the third class of words that will follow more or less the
same frequency distribution 
in the text and in the vocabulary are the {\it common words}. 
They serve to transmit the meaning of the text, but are common for every text
that must explain some concept,
for example, like the word ``explain'' in this sentence.

The paper is organized in the following way: In 
Section 2 we define the model 
and the approximations used. 
In Section 3 we calculate numerically the
thermodynamic quantities for a set of arbitrary selected texts.
We show that the specific terms (keywords) and the rest of the text have different 
thermodynamic behavior. 
Finally in Section 4 we present our conclusions.

%% file: method_tr.tex
\section{The Model}
The relationship between some text and its language can be considered as a conditional distribution. 
In this article however we consider the following approach:

We consider the vocabulary as a solid state basement, composed by ``molecules'', which are actually the 
parts of the text (the words of the language).
The text itself is considered as a liquid solution of ``molecules'', derived in the same manner as the vocabulary.
The text and the vocabulary ``react'' and there exists some energy gain when the reaction takes place, 
so some ``molecules'' are settled down on the solid base.

The excess of the ``molecules''
(words) of a given type in the vocabulary,
e.g. in the ``solid'' compound, has no significant meaning.
Therefore, we can concentrate only on the ``liquid'' phase,
considering this phase as a significant one. 
Equivalently, we can consider only on the deposited part of the
``molecules'' that have been entered in 
reaction with the vocabulary. 
The molecules as a first approximation can be assumed to react only if
they represent one and the same word.

More rigorously, our model consists of a vocabulary with length $L_v$, a text
of length $L_t$, the ``molecules'' 
(words) of the text, $w$, that are matched with the ``molecules'' of the
vocabulary and the corresponding number of 
occurrences of these ``molecules'': $n_t(w)$ for the text and $n_v(w)$ for the vocabulary. 
In order to fulfill the requirement of equal molar mass we can
introduce some standard text length $L_0$ and 
normalize the number of occurrence of $w$ according to this length:
\[
N_t(w)=L_0\frac{n_t(w)}{L_t},\ \ \ N_v(w)=L_0\frac{n_v(w)}{L_v}.
\]
For convenience we choose $L_0=L_t$ in the numerical experiments.
We denote by $m(w)$ the number of deposited molecules, normalized to a length $L_0$. This parameter will be used below as an order parameter for
  the system.

The problem of regarding the text as a thermodynamic system 
consists of defining the ``molecules'' $w$ and 
the energy of interaction 
$E(w)=E(m(w),N_t(w),N_v(w),L_0)$
between them.

In this article we will regard the ``molecules'' as usual English
words, consisting of continuous strings of 
letters, separated by non-letter symbols. 
In the rest of the article, we will not distinguish between ``molecules'' and words.
We will assume that the words are independent, e.g. that there is no
interaction between the different words.
Due to this independence, the extensive thermodynamics quantities, as for example the free energy, 
will be given by the sum of the individual quantities corresponding to the different words. 
Therefore we can build a theory, based on a single word, and extrapolate it on the whole text.

Further, we will consider that the language (the solid compound) imposes some potential energy field 
depending on the parameters $N_v$, $L_v$, but not on the text,
e.g. not on $N_t$.
We also assume that the system is in thermal equilibrium. 

According to the general thermodynamics principles, the state of the system
can be described only by its energy $E$. The probability $P(m)$ of the
state with $m$ deposited molecules (words) is:
\begin{equation}
P(m) \propto G(m) \exp(- \beta E(m, N_t, N_v, L_0)),\label{pw}
\end{equation}
where $E(m, N_t,N_v, L_0)$ is the energy of settling $m$ molecules,
$G(m)$ is the number of degeneration of 
these states and $\beta$ is the inverse temperature $\beta\equiv1/T$.

The number of degeneration is just the number of ways one can select $m$ molecules
out of a set of $N_t$ molecules, e.g. ${{N_t}\choose {m}}$. 
Note that this number is strictly zero if $m>N_t$ due to the fact that we have only $N_t$ molecules.

Regarding that system, one can impose the requirement that its thermodynamic
properties scale with the length of the texts, e.g. 
if we scale simultaneously the 
size of the vocabulary and the size of the text by $s$, then
the thermodynamic potentials scale as:
\[
E(s m, s N_t, s N_v, s L_0)= s E(m, N_t, N_v, L_0)
\]
and
\[
\log(G(s m))= s \log(G(m)).
\]
Considering the frequency of occurrence of a single word $w$ in a 
text with length $L$, in order to speak about something measurable, we
must regard the case where the word occurs in the text $x\gg 1$ times.

Let us assume that we have a language derived from some context free weighted 
grammar 
with terminal symbols --- the words of the language.
Being interested only on the frequency of the words, we can transform any 
language definition $A: \alpha w \gamma$ into $A: w \alpha \gamma$.
Further, all words different from $w$ can be regarded as one and the same 
word, say $v$, thus obtaining a grammar with only two terminal symbols. 
We can join further in one symbol all non-terminal symbols that can not produce $w$.

The simplest grammar of this type is:

\begin{eqnarray} 
S:& w R  & \nonumber\\
R:& v R  & [p] \nonumber\\
R:& \lambda  & [1-p] , \nonumber\\
\end{eqnarray}
where $S$ is the axiom, $R$ is an auxiliary non-terminal symbol and 
$\lambda$ is en empty string. Within the square brackets we represented 
the probabilities for the corresponding definitions.
This grammar generates sentences containing $w$ with lengths that have 
exponentially falling  frequencies distributions.

Fixing the length of the text, we can obtain the probability distribution
of $w$ as a sum of 
finite number of exponentially distributed variables.
By definition \cite{specfunc}, this is the gamma distribution:
\begin{equation}\label{gamamdist}
P(x;w)= e^{-x b}x^{a-1}b^a /\Gamma(a).
\end{equation}
Here the parameter $a$ is proportional to the length of the text $L$, while the parameter $b$ does not depend on it, but rather on the word and the class of text we are regarding.
Although the above consideration is not strict, 
it can give an idea about the type of the word frequency distribution for a 
text with a fixed length. 
Using a general context free grammar, one can find the expressions for the 
probability and the length of the sentence, following the techniques developed in \cite{c1}.

We have checked the hypothesis for the gamma distribution 
on a set of about 19000 English 
texts given by the Gutenberg collection and have found an excellent agreement with the experimental 
data for all the words with frequency of occurrence $p(w)>5/10000$.

\begin{figure}[th]
\begin{center}
\begin{minipage}{6cm}
\epsfxsize 6cm 
\epsfbox{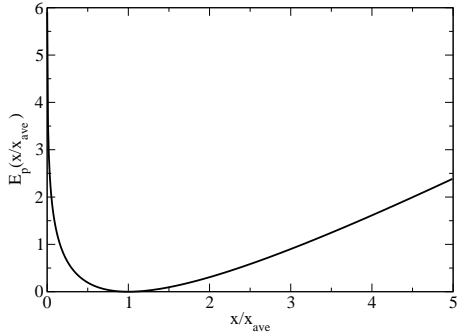}
\end{minipage}
\caption{\small
Potential energy of a word according to the number of words. It consists of 
two parts -- 
the logarithmic falling part, varying for values of the argument from zero 
to the mean frequency of the word,
and a 
linear increasing part, predominant at the range where the frequency of the 
word is larger
than its mean frequency.
}\label{potenergy}
\end{center}
\end{figure}

Later we have studies the asymptotic behavior of the distribution. 
For this aim, we have replicated the text $s$ times and have considered the limit 
$\lim_{s\rightarrow\infty}[\log P(s x; w; s a,b)]/s$. 
One can easily find that this limit is $a-b x-a \log a + a \log x + a \log b$.
Using that the mean of $x$ is $\bar x=a b$, we finally obtained the following expression  for the asymptotic behavior of $P(x)$:
\begin{equation}\label{Ep}
E_p(x;w)=\log P(x)= \bar x b \left[ 1-\frac{x}{\bar x}+
\log\left(\frac{x}{\bar x}\right)\right].
\end{equation}
$E_p$ can be regarded as a potential energy of the word $w$ in the language. 
The linear member accounts for the excess of words of a given type in the 
text, while the logarithmic one corresponds to the entropic part of the energy \cite{chaos}.
A typical energy curve is given in Fig.~\ref{potenergy}. 
\vskip3mm
Using the above considerations, the corresponding partition
function for a given word $w$ is:
\begin{equation}\label{Partfunc}
Z(w, \beta)=\sum_{m=1}^{N_t} G(m, N_t)) \exp(-\beta E_p(m, N_t)),
\end{equation}
where $G(m, N_t) = {{N_t}\choose {m}}$ and we have omitted
the argument $w$ in the right hand side of the equation.
Thus, the expression for the partition function is:
\begin{equation}\label{z1}
Z(w, \beta) = \sum_{m=1}^{N_t} \exp(- \beta E_{tot}(m, N_t)),
\end{equation}
where
\begin{eqnarray}\label{z2}
E_{tot}(m, N_t)= - \frac{1}{\beta} \log {{N_t}\choose {m}} + \nonumber\\
N_v b \left[ 1-\frac{m}{N_v}+
\log\left(\frac{m}{N_v}\right)\right]
\end{eqnarray}
is the total energy corresponding to a given word $w$. It is composed by a 
potential part $E_p$ and a combinatorial part 
$\frac{1}{\beta}\log G(m, N_t)$.

Finally, the full free energy of the text is given by the sum 
over the different words of the text:
\begin{equation}
F(\beta) = - \frac{1}{\beta} \sum_{w} \log Z(w,\beta).
\end{equation}
The equation for the order parameter $m$ can be obtained by using the 
saddle-point method. 
Its application, combined with the 
Stirling approximation, $\log N! \approx N \log N - N$, $N\gg 1$,
gives the following equation for the order parameter $m$:
\begin{equation}\label{m1}
\frac{d E_{tot}}{dm} = \frac{1}{\beta}\log{\frac{m}{N_t-m}}+b\frac{N_v-m}{m}=0.
\end{equation}
This equation can be solved in closed form and the solution is:
\begin{equation}\label{m2}
m = N_t\frac{b \beta N_v/N_t}
           {b \beta N_v /N_t
             + W(b \beta N_v/N_t\ e^{b \beta - b \beta N_v/N_t})},
\end{equation}
where $W(.)$ is the Lambert W function \cite{specfunc}.

The entropy $S$ for a single word is:
\begin{eqnarray}\label{entrop}
S \equiv - \frac{\partial F}{\partial T} = N_t \log N_t -m \log m - 
\nonumber\\
(N_t-m) \log (N_t - m).
\end{eqnarray}

\begin{figure}[h]
\begin{center}
\begin{minipage}{7cm}
\epsfxsize 7cm 
\epsfbox{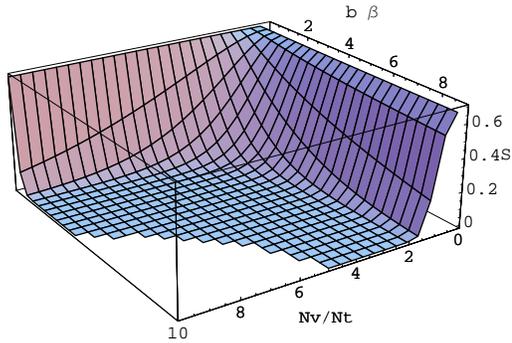}
\end{minipage}
\caption{\small
The entropy for a single word.
}\label{entropy}
\end{center}
\end{figure}

Substituting Eq. (\ref{m2}) in Eq. (\ref{entrop}),
we obtained the behavior of the entropy as a function of the inverse 
temperature and the ratio $N_v/N_t$ shown in Fig.~\ref{entropy}. 

\begin{figure}[h]
\begin{center}
\begin{minipage}{7cm}
\epsfxsize 7cm 
\epsfbox{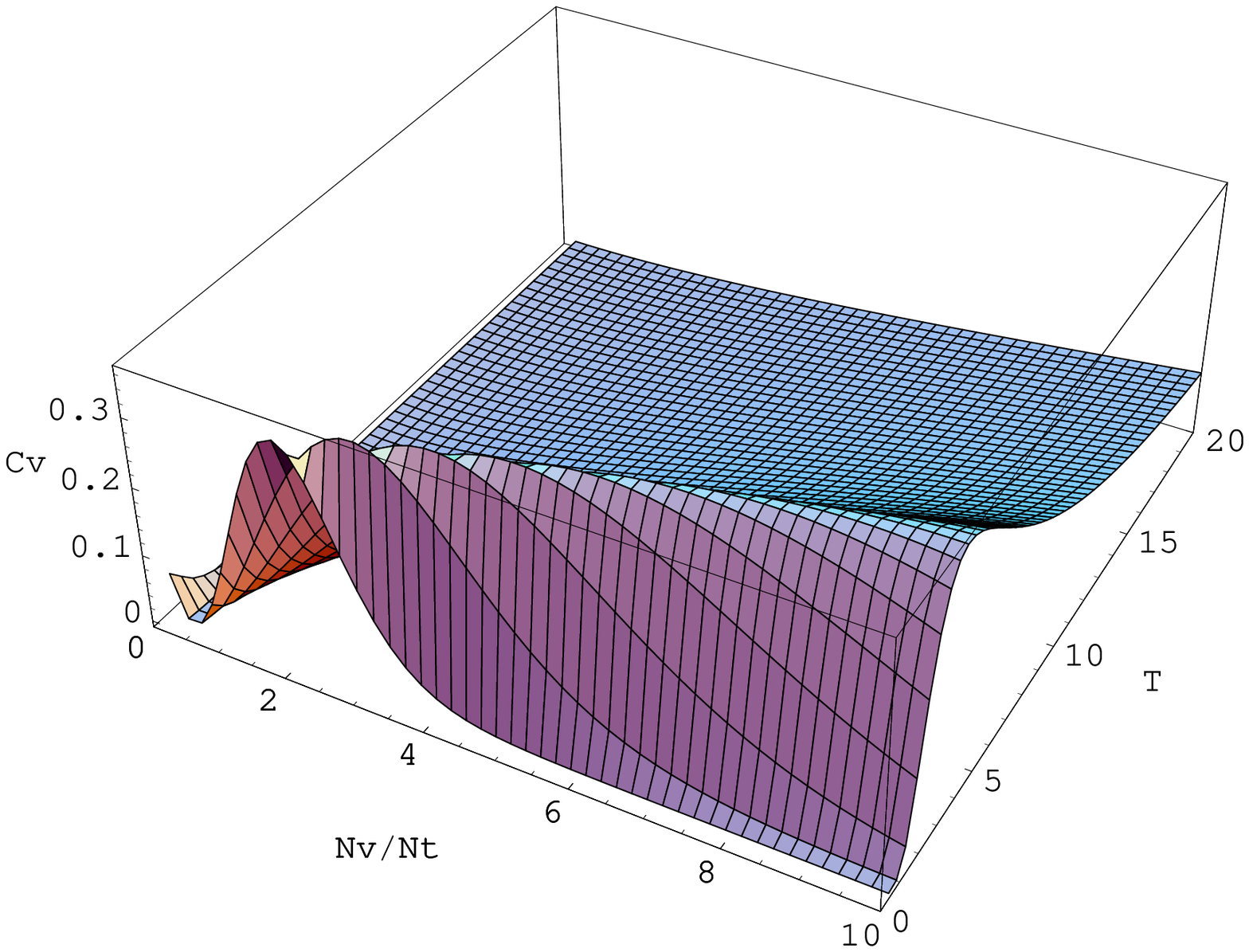}
\end{minipage}
\hfill
\begin{minipage}{6cm}
\epsfxsize 6cm 
\epsfbox{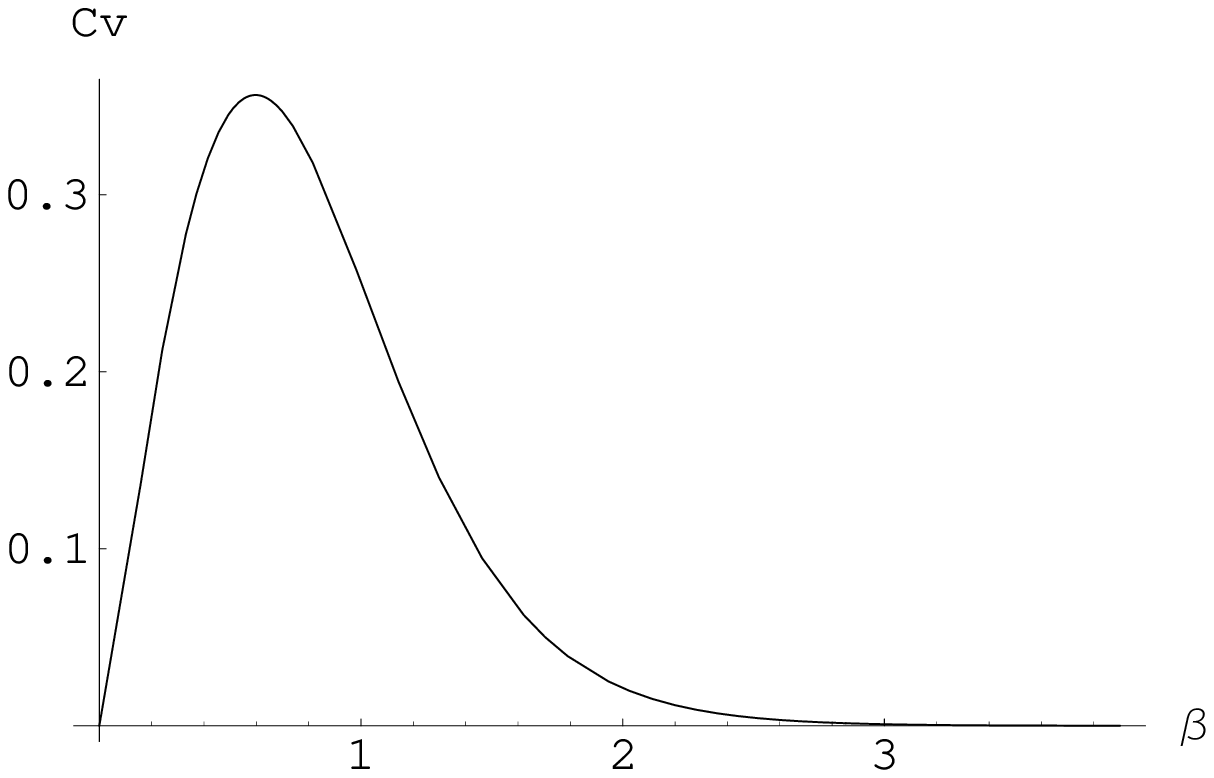}
\end{minipage}
\caption{\small
The ``specific heat'' $C_V$.
}\label{figcv}
\end{center}
\end{figure}

Finally, the second derivative of the free energy, which is related to the
``specific heat'', 
\[
C_V = -T \left(\frac{\partial ^2 F}{\partial T^2}\right)_V,
\] 
is represented in Fig.~\ref{figcv}. We have used the notation adopted in 
the thermodynamics for isochoric process, $C_V$, 
although what is fixed in the present approach is
the number of occurrences  for a given word.
A  section of that figure 
for $b=1, N_v = 5$ is represented in the right panel of the same figure.

%% file: numerical_tr.tex
\section{Numerical experiments}

To check the above results experimentally on real texts, we used a collection of about 19000 English texts from the Gutenberg project (GC)
with size of $5.10^7$ words (GC). We also used a 
collection of 500 articles from the area of the non-linear physics (NL) given by the repository xxx.archiv.gov.
In order to avoid problems with the different multiple versions of the articles, we used only the 
first version of each one.
Finally, we used a list of 257 closed-class words of English instead of function words.

For estimating the parameters $a$ and $b$ of the distribution of a single 
word,
we used GC.
We found that $b$ is within the range [0.01-20] with an average value 0.25. The value of the parameter $a$, for a text with a length $L=10000$, is within the range [0-2.6].
Note that the parameters $a$ and $b$ are well defined and with a sufficient 
confidence only if $p(w) L\gg 1$. For practical purposes we can say that
within a corpus of $10^8$ words, 
these two parameters are well defined only for the 2400 most frequent words.
For the rest of the words, we used some 
simplifying assumption, due to the fact that one can not prove or disprove
reliably the hypothesis by using two degrees of freedom ($a$ and $b$), having less then four measures for 
their estimation.
The hypothesis, we have adopted, was that 
the less frequent words have the same value for the
parameter $b$ of the distribution. Thus we can join all the words, that are 
not frequent enough, and estimate that parameter. 
The results are very close to the mean vale of $b$. The parameter $a$, being proportional to the
length of the text, is not so critical for the estimation (actually we need only $N_v$ and $b$). 

\begin{figure}[h]
\begin{center}
\begin{minipage}{5.70cm}
\epsfxsize 5.7cm 
\epsfbox{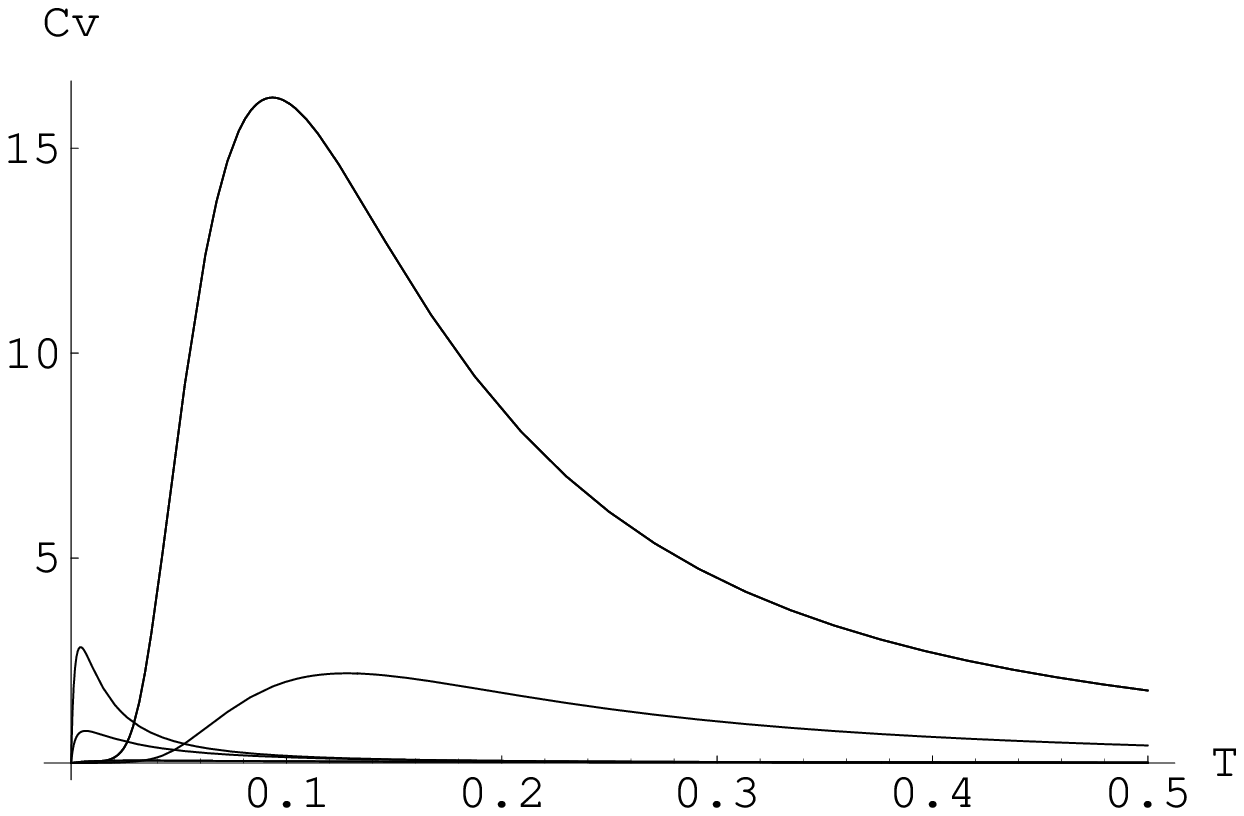}
\end{minipage}
\hfill
\begin{minipage}{5.70cm}
\epsfxsize 5.7cm 
\epsfbox{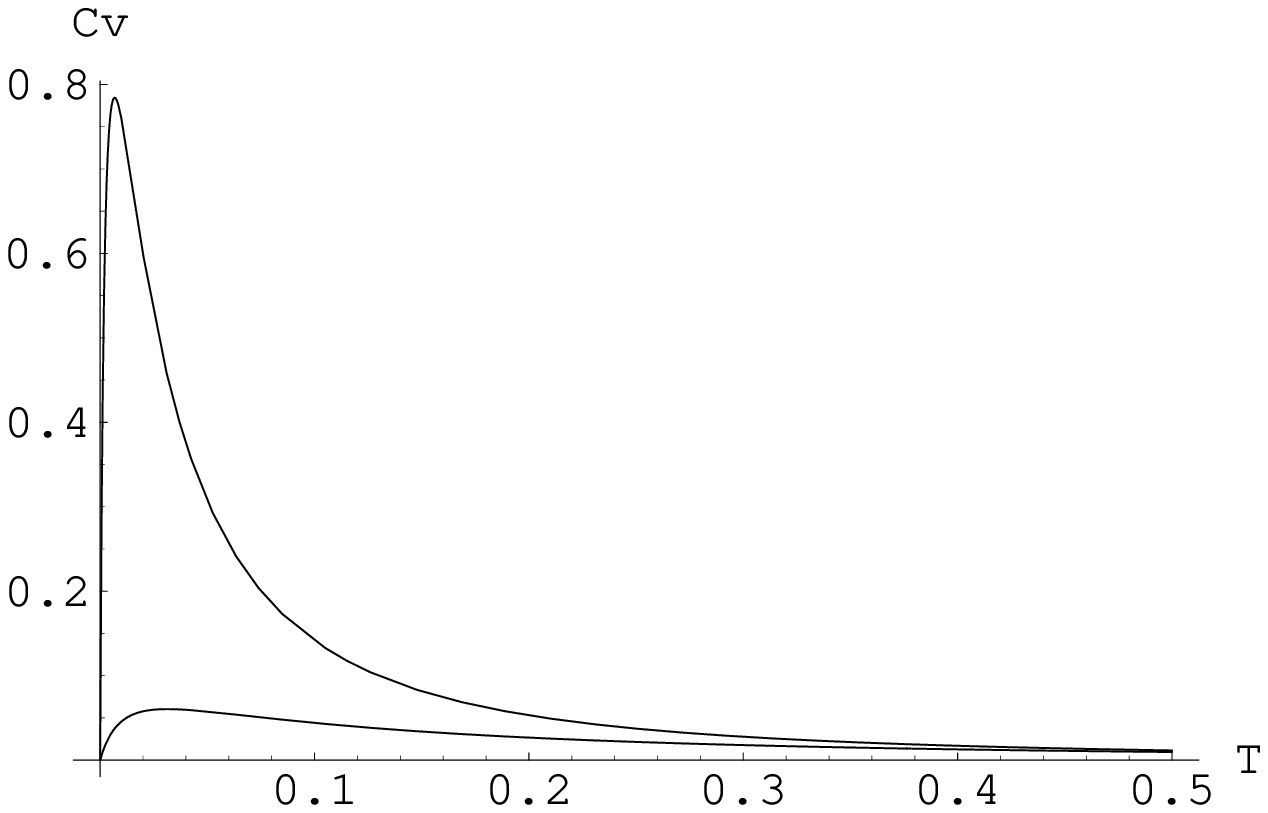}
\end{minipage}
\caption{\small
The specific heat $C_V$ for different words belonging to one and the same text. The upper two curves in 
the left panel represent two different terms (``topology'' and 
``topological''). The lower curves of the left panel 
represent two function words 
(``the'' and ``are''). On the right panel, the curve of the word ``are'' is 
zoomed in order to represent it together with the typical common word ``important''.
}\label{cvt}
\end{center}
\end{figure}

Figs.~\ref{cvt} 
show the typical behavior of the specific heat $C_V$ for different kind of words: for terms (the two upper curves on the left
panel),  for function words (the two lower curves of the same panel) 
and 
for common words (the lower curve on the right panel). 
One can observe that $C_V(N_t,N_v,b)$ represents 
different behavior for the different word classes 
with corresponding maxima belonging to
different temperature ranges.

Because the function words have a higher frequency of occurrence, one can expect that
they play a predominant role for the behavior of the specific heat.
As we can observe, this is not true:
the specific heat for the terms is much higher 
than the one corresponding to the function words.
These results can be interpreted as an indication than the most vulnerable 
parts of the speech are carried by the common words, while the most resistant ones are carried by the
domain-specific terms.

%% file: conclusion_tr.tex
\section{Conclusion}

In the present article we proposes a statistical mechanics approach for the
analysis of the human written text. By introducing an energy,
that describes the system, taking into consideration a realistic 
distribution of the 
words inside a large text corpus, we were able to 
derive the thermodynamic parameters of the system 
in a closed analytical form.

By studying the behavior of the specific heat of the system, we have shown
that this quantity is different for different kinds of words 
(terms, function words and common words).

We have applied the above method to different
corpora of texts and we have found one and the same universal behavior, which does not depend on the particular text.

Our numerical results show that 
the ``specific heat''  
effectively separates the closed class words 
from the specific terms and the common words used in the text.